\documentclass[sigconf]{acmart}
\AtBeginDocument{%
  \providecommand\BibTeX{{%
    \normalfont B\kern-0.5em{\scshape i\kern-0.25em b}\kern-0.8em\TeX}}}

\copyrightyear{2022}
\acmYear{2022}
\setcopyright{rightsretained}

\acmConference[SIGIR '22]{Proceedings of the 45th International ACM SIGIR Conference on Research and Development in Information Retrieval}{July 11--15, 2022}{Madrid, Spain}
\acmBooktitle{Proceedings of the 45th International ACM SIGIR Conference on Research and Development in Information Retrieval (SIGIR '22), July 11--15, 2022, Madrid, Spain}
\acmDOI{10.1145/3477495.3532069}
\acmISBN{978-1-4503-8732-3/22/07}

\usepackage{color}
\usepackage{multirow}
\usepackage{multicol}

\usepackage{adjustbox}

\usepackage{etoolbox}
\makeatletter
\patchcmd{\maketitle}{\@copyrightpermission}{
  \begin{minipage}{0.3\columnwidth}
     \href{http://creativecommons.org/licenses/by/4.0/}{\includegraphics[width=0.90\textwidth]{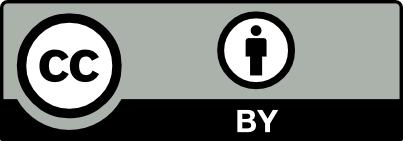}}
  \end{minipage}\hfill
  \begin{minipage}{0.7\columnwidth}
     \href{http://creativecommons.org/licenses/by/4.0/}{This work is licensed under a Creative Commons Attribution International 4.0 License.}
  \end{minipage}
  
  \vspace{5pt}
}{}{}

\makeatother

\begin{document}

\fancyhead{}

\title{SPACE-3: Unified Dialog Model Pre-training for Task-Oriented Dialog Understanding and Generation}

\author{Wanwei He}
\authornote{Equal Contribution.}
\authornote{Wanwei He is also with the University of Chinese Academy of Sciences. This work was conducted when Wanwei He was interning at Alibaba.}
\affiliation{%
  \institution{Shenzhen Institute of Advanced Technology, Chinese Academy of Sciences}
  \city{Shenzhen}
  \country{China}}
\email{ww.he@siat.ac.cn}

\author{Yinpei Dai}
\authornotemark[1]
\affiliation{%
  \institution{Alibaba Group}
  \city{Beijing}
  \country{China}}
\email{yinpei.dyp@alibaba-inc.com}

\author{Min Yang}
\authornote{Corresponding authors.}
\affiliation{%
  \institution{Shenzhen Institute of Advanced Technology, Chinese Academy of Sciences}
  \city{Shenzhen}
  \country{China}}
\email{min.yang@siat.ac.cn}

\author{Jian Sun}
\affiliation{%
  \institution{Alibaba Group}
  \city{Beijing}
  \country{China}}
\email{jian.sun@alibaba-inc.com}

\author{Fei Huang}
\affiliation{%
  \institution{Alibaba Group}
  \city{Beijing}
  \country{China}}
\email{f.huang@alibaba-inc.com}

\author{Luo Si}
\affiliation{%
  \institution{Alibaba Group}
  \city{Beijing}
  \country{China}}
\email{luo.si@alibaba-inc.com}

\author{Yongbin Li}
\authornotemark[3]
\affiliation{%
  \institution{Alibaba Group}
  \city{Beijing}
  \country{China}}
\email{shuide.lyb@alibaba-inc.com}

\begin{abstract}
Recently, pre-training methods have shown remarkable success in task-oriented dialog (TOD) systems. However, most existing pre-trained models for TOD focus on either dialog understanding or dialog generation, but not both.
In this paper, we propose SPACE-3, a novel unified semi-supervised pre-trained conversation model learning from large-scale dialog corpora with limited annotations, which can be effectively fine-tuned on a wide range of downstream dialog tasks.
Specifically, SPACE-3 consists of four successive components in a single transformer to maintain a task-flow in TOD systems: (i) a dialog encoding module to encode dialog history, (ii) a dialog understanding module to extract semantic vectors from either user queries or system responses, (iii) a dialog policy module to generate a policy vector that contains high-level semantics of the response, and (iv) a dialog generation module to produce appropriate responses. We design a dedicated pre-training objective for each component. Concretely, we pre-train the dialog encoding module with span mask language modeling to learn contextualized dialog information. To capture the structured dialog semantics, we pre-train the dialog understanding module via a novel tree-induced semi-supervised contrastive learning objective with the help of extra dialog annotations.
In addition, we pre-train the dialog policy module by minimizing the $\mathcal{L}_2$ distance between its output policy vector and the semantic vector of the response for policy optimization.
Finally, the dialog generation model is pre-trained by language modeling.
Results show that SPACE-3 achieves state-of-the-art performance on eight downstream dialog benchmarks, including intent prediction, dialog state tracking, and end-to-end dialog modeling.
We also show that SPACE-3 has a stronger few-shot ability than existing models under the low-resource setting.\footnote{For reproducibility, we release our code, pre-training data and pre-trained model publicly at \url{https://github.com/AlibabaResearch/DAMO-ConvAI/tree/main/space-3}.}
\end{abstract}



\maketitle

\begin{figure}
    \centering
    \includegraphics[width=0.4\textwidth]{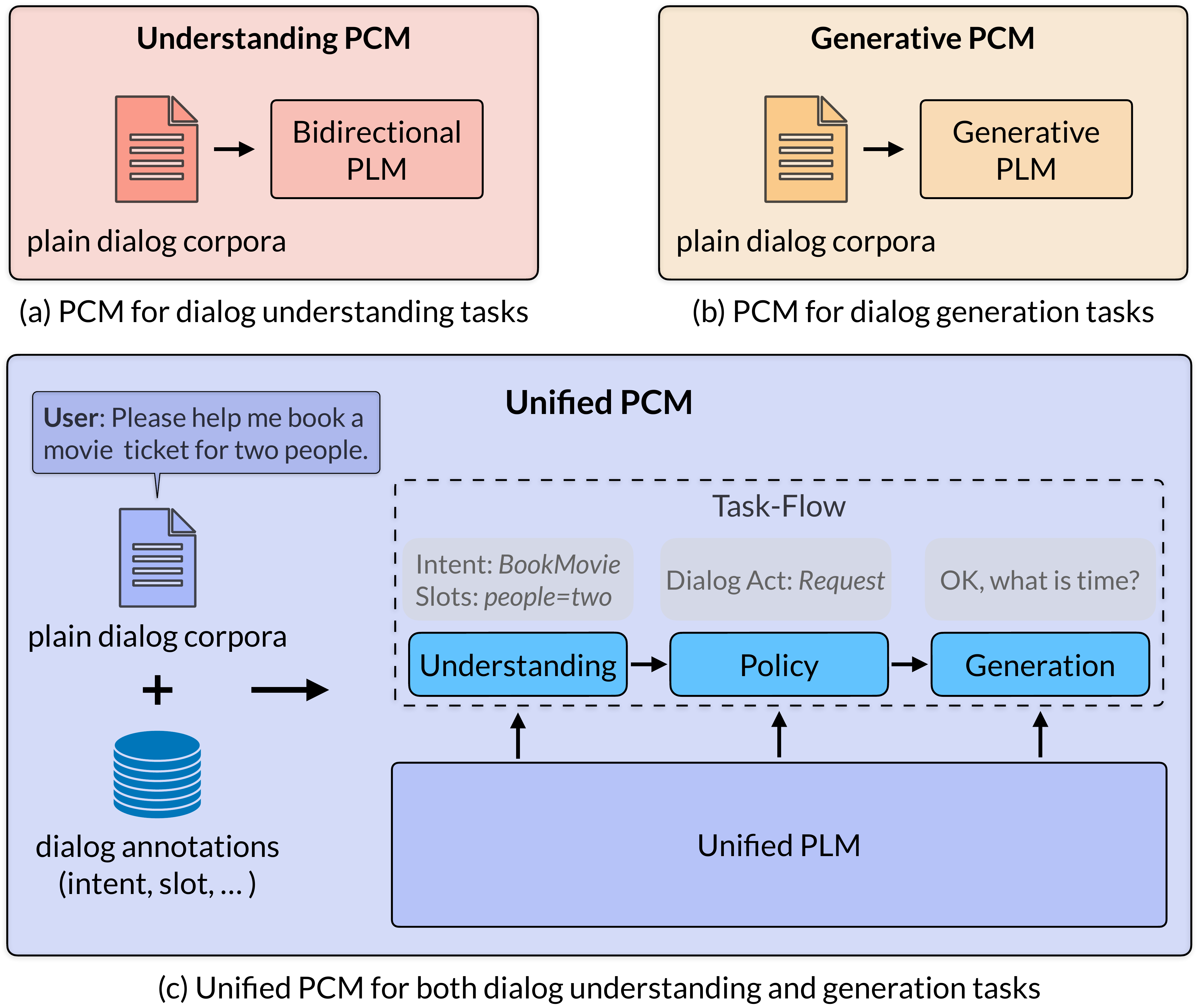}
    \caption{
    (a) The pre-trained conversation model (PCM) for dialog understanding, which directly pre-trains bidirectional language models on plain dialog corpora.
    (b) The PCM for dialog generation, which employs a generative PLM on plain dialog corpora.
    (c) We propose a unified PCM for both dialog understanding and dialog generation, which maintains the task-flow and is trained on both labeled and unlabeled data.}
    \label{fig:intro}
\end{figure}

\section{Introduction}
Task-oriented dialog (TOD) systems, which aim at helping users accomplish specific tasks through natural language interactions, have attracted increasing attention recently due to the broad applications such as flight booking, weather inquiry and restaurant reservation.
To fulfill user goals, TOD systems must be capable of understanding the dialog, planning the policy, and generating human-like responses, which maintain a task-flow during the dialog interaction. Concretely, as illustrated in Figure \ref{fig:intro} (c), a complete task-flow consists of three phases: (i) extracting crucial semantics of user query such as intents and slots, conditioned on dialog context (\textit{Dialog Understanding}); (ii) determining dialog acts towards successful task completion (\textit{Policy Planning}); (iii) generating natural and appropriate responses (\textit{Dialog Generation}).


Recent advances of pre-trained language models (PLMs) \cite{devlin-etal-2019-bert, liu2019roberta, yang2019xlnet, radford2019language} have brought significant improvement
by further fine-tuning on downstream various downstream tasks \cite{mehri2020dialoglue, zhang2019find, heck-etal-2020-trippy}.
However, previous studies \cite{zhang2019dialogpt, wu-etal-2020-tod} reveal that there are intrinsic  differences in linguistic patterns between human conversations and plain texts, which leads to deficiencies of TOD systems.
Subsequently, many methods attempt to address this issue by building pre-trained conversation models (PCMs) \cite{mehri2020dialoglue, zhang2019dialogpt, wu-etal-2020-tod}, which further pre-train vanilla PLMs on large-scale dialog corpora to capture conversational knowledge.
For dialog understanding tasks (e.g., intent recognition and dialog state tracking), ConvBERT \cite{mehri2020dialoglue} continually pre-trained a BERT model on large-scale human-human conversational logs with the masked language modeling loss. TOD-BERT \cite{wu-etal-2020-tod} further integrated a contrastive objective with a  better response selection mechanism.
For dialog generation tasks, 
DialoGPT \cite{zhang2019dialogpt} was the first to continually pre-train a GPT-2 on 147M conversations from Reddit to generate more relevant and consistent responses than strong RNN baselines.
PLATO \cite{bao-etal-2020-plato} leveraged both Twitter and Reddit data to pre-train a dialog generation model with discrete latent variables, which implicitly modeled dialog policy and alleviated the one-to-many mapping problem in open-domain dialog generation.

Despite the remarkable progress of previous PCMs on dialog understanding or dialog generation, there are still several technical challenges to constructing an effective and unified pre-trained conversation model. \textbf{First}, most current PCMs are tailored for one specific task like dialog understanding or dialog generation, as illustrated in Figure \ref{fig:intro} (a)-(b). Limited exploration has been attempted to solve the three sub-tasks jointly in a unified framework. For example, although TOD-BERT significantly improves the performance of a wide range of dialog understanding tasks \cite{wu-etal-2020-tod},  it is difficult to apply TOD-BERT for dialog generation due to its bidirectional nature \cite{wang-cho-2019-bert}. 
\textbf{Second}, a common idea of existing PCMs is to directly train existing PLMs on dialog corpora with vanilla language model objectives while neglecting the task-flow characteristics in TOD. For example, PPTOD~\cite{su2021multitask} proposes to pre-train a dialog model by primitively amalgamating different dialog tasks with heterogeneous annotations into a text-to-text format like T5, which decreases the performance on individual TOD tasks.
Different from open-domain chatting bots, TOD systems aim to help users accomplish certain tasks with a controllable step-by-step procedure, therefore it is essential to explicitly incorporate the task-flow into pre-training for fully exploiting task-oriented dialog features.  
\textbf{Third}, in previous PCMs, leveraging semantic structures and manual annotations (e.g., intents or slots) of dialogs  to learn better pre-trained dialog representations still remains unexplored. Nevertheless, there have been many works demonstrating that labeled data can not only accelerate the pre-training procedure but also improve the model performance \cite{assran2020supervision, khosla2020supervised, dai2021unimoco}. Therefore, we argue that it is  also crucial to incorporate dialog annotations and semantic structures to learn powerful PCMs.

In this paper, all the aforementioned challenges are considered in a holistic framework. We propose a novel end-to-end semi-supervised pre-trained conversation model (called SPACE-3), which can be effectively applied to various downstream dialog tasks.
SPACE-3 is pre-trained on both labeled and unlabeled dialog data, aiming to inject the knowledge of dialog annotations into task-oriented dialog pre-training on large-scale task-oriented dialog corpora.
To be specific, SPACE-3 consists of four successive core components to model a complete task-flow in a unified transformer architecture, including (i) a dialog encoding module to encode dialog history and capture common context representations, (ii) a dialog understanding module to extract semantic vectors from either user queries or system responses, (iii) a dialog policy module to learn a policy vector  that contains high-level semantics of the response, and (iv) a dialog generation module to produce appropriate responses. Except the encoding module is bidirectional, all the rest three modules are set as unidirectional to preserve the causal relations within the task-flow. We also design a dedicated pre-training objective for each component. Concretely, we pre-train the dialog encoding module with span mask language modeling to learn the contextualized dialog information. To capture semantic dialog structure knowledge, we pre-train the dialog understanding module via a novel tree-induced semi-supervised contrastive learning objective with the help of extra dialog annotations. We treat structured semantic annotations stored in labeled dialog corpora as trees, and estimate the similarity of different dialogs by Tree Editing Distance.
According to the soft similarity scores, supervised contrastive learning is applied to increase the similarity of semantically similar dialogs and push away the representations of all other dialogs. We also adopt vanilla self-supervised contrastive learning to refine the learned representation with the help of unlabeled dialogs.
In addition, the dialog policy module is pre-trained to determine the semantics of the system response by minimizing the $\mathcal{L}_2$ distance between its output policy vector and the semantic vector of the response extracted by the dialog understanding module.
Finally, the dialog generation model is pre-trained by language modeling. 
We aggregate all pre-training objectives above to learn  dialog understanding, policy planning, and dialog generation abilities from both limited labeled dialog datasets and large-scale unlabeled dialog corpora.   
It is noteworthy that SPACE-3 can avoid error accumulation due to the end-to-end pre-training.

We summarize our main contributions as follows:
\begin{itemize}
    \item To the best of our knowledge, we are the first to propose a unified PCM for dialog understanding, policy planning, and dialog generation, which fully exploits task-oriented dialog characteristics and maintains the task-flow by connecting the successive dialog-related tasks.
    \item We explore tree-induced semi-supervised contrastive pre-training for dialog understanding. The injection of structured semantic annotations via contrastive learning enhances dialog representation learning on large-scale dialog corpora.
    \item For pre-training SPACE-3, we construct a labeled corpus \textit{AnPreDial} with 3 million utterances and a large-scale unlabeled dialog corpus \textit{UnPreDial} with 19 million utterances.  
    Experiments show that our pre-trained model achieves new state-of-the-art performance on eight downstream dialog benchmarks, including intent prediction, dialog state tracking, and end-to-end dialog modeling.
    We believe that the release of the constructed dialog corpora and the unified PCM would push forward the research in this area.  
\end{itemize}

\section{Related Work}
\subsection{Pre-trained Language Models}
Pre-trained language models (PLMs) can be roughly divided these PLMs into two categories (uni-directional and bi-directional) according to the attention mechanisms applied in the pre-training.
For natural language understanding tasks like classification \cite{wang-etal-2018-glue} or machine reading comprehesion \cite{rajpurkar-etal-2016-squad}, BERT \cite{devlin-etal-2019-bert} and RoBERTa \cite{liu2019roberta} were pre-trained with a bi-directional transformer, aiming to grasp deep contextualized semantic meanings.
For natural language generation tasks, GPT \cite{radford2018improving, radford2019language} and T5 \cite{JMLR:v21:20-074} employed the uni-directional transformer encoder to maximize left-to-right generation likelihood. 
Recently, the unified language model UniLM \cite{dong2019unified} enabled both uni-directional and bi-directional attention with flexible self-attention mask designs.
However, previous works \cite{zhang2019dialogpt, kulhanek2021augpt} reveal that there are intrinsic differences between plain texts and human conversations.
Directly fine-tuning PLMs, which are pre-trained on plain texts, on downstream dialog tasks prevents the model from sufficiently capturing conversational patterns and linguistic knowledge, which leads to sub-optimal results \cite{mehri2019pretraining, zenginvestigation, wu2020probing}.

\subsection{Pre-trained Conversation Models}
To bridge the gap between plain texts and human conversations, many studies attempted to build pre-trained conversation models (PCMs) by further pre-training PLMs on dialog corpora \cite{mehri2020dialoglue, zhang2019dialogpt, henderson2019convert}, which achieved impressive performance on both dialog understanding \cite{wu-etal-2020-tod} and generation tasks \cite{peng2020few}.
A majority of PCMs are trained on open-domain conversational data from Reddit or Twitter for dialog response generation.
For example, DialoGPT \cite{zhang2019dialogpt} extended and pre-trained the GPT-2 on 147M conversations extracted from Reddit. Blender \cite{roller2020recipes} was pre-trained on 1.5B open-domain dialogs, demonstrating powerful dialog generation performance.
Furthermore, 
PLATO \cite{bao-etal-2020-plato} used discrete latent variables to tackle the one-to-many mapping problem in open-domain dialog generation.

Different from open-domain dialog systems, task-oriented dialog (TOD) systems have explicit goals (e.g. flight booking or restaurant reservation), making dialog understanding and policy planning components important \cite{wu-etal-2020-tod,he-etal-2020-amalgamating,HE2021106667}.
The PCMs for TOD systems can be roughly divided into two types.
The first type is designing PCMs that are tailored for dialog understanding tasks.
For example, ConvBERT \cite{mehri2020dialoglue} utilized a masked language modeling objective to fine-tune BERT on a large dialog corpus so that semantically rich representations can be learned.
TOD-BERT \cite{wu-etal-2020-tod} employed a contrastive objective for response selection and incorporated user/system tokens into the masked language modeling.
DialogueBERT \cite{zhang2021dialoguebert} explored the masked utterance modeling and response contrastive loss, showing good performance on intent, emotion, and entity recognition.
The second line is to design PCMs that are tailored for dialog generation tasks.
SC-GPT \cite{peng2020few} focused on the natural language generation (NLG) task, assuming that dialog acts and slot-tagging results are given to generate a natural language response.
SOLOIST \cite{TACL2877} parameterized a task bot using a Transformer-based auto-regressive language model, which subsumed different dialog modules into a single neural model and was pre-trained on two TOD datasets.
SPACE \cite{he2022galaxy} proposed to use consistency regularization loss to learn dialog policy from labeled and unlabeled dialog corpora via a semi-supervised manner. 
To exploit more heterogeneous TOD corpora, PPTOD  \cite{su2021multitask} converted different TOD tasks into the text-to-text generation task with task-specific prompts on T5 model \cite{JMLR:v21:20-074}.
Different from these methods, we propose SPACE-3 that explicitly models task-flow during pre-training while avoiding the error accumulation in the pipeline TOD architecture via multi-task training.

\subsection{Contrastive Learning}
Contrastive learning (CL) is a popular pre-training method, which takes advantage of self-supervised objectives on large-scale plain texts to learn semantic representations.
The key idea of contrastive learning is to narrow the distance between two semantically similar sentence representations and push away the representations of dissimilar sentences \cite{gao2021simcse, wu2021esimcse, yan2021consert}.
Recently, contrastive learning has also attracted increasing attention for dialog understanding tasks.
For example, \cite{zhang2021few} performed self-supervised pre-training by employing contrastive learning objectives and fine-tuned the pre-trained models on few-shot intent recognition tasks.
\cite{he2020contrastive} used adversarial samples as difficult negative instances in contrastive learning for zero-shot cross-domain slot-filling.
The most related works to ours are \cite{mehri-eric-2021-example, vulic2021convfit}, which utilized supervised contrastive learning at the fine-tuning stages for intent recognition tasks, where the samples from the same class are all regarded as positive ones. Different from \cite{mehri-eric-2021-example, vulic2021convfit}, we explore semantic structures in TOD, and propose a new contrastive learning approach that calculates semantic tree-structure similarity among all possible labeled data, which is essential for the semantic modeling in the task-flow.

Subsequently, semi-supervised contrastive learning has been proven as a promising pre-training approach in various research fields, such as image recognition \cite{yuan2021activematch, park2021opencos, li2021comatch, kim2021selfmatch}, image segmentation \cite{alonso2021semi, zhou2021c3} and speech recognition \cite{xiao2021contrastive, inoue2020semi}.
It adopts supervised and self-supervised contrastive objectives together on the combination of labeled and unlabeled data.
However, semi-supervised contrastive learning  remains unexplored in task-oriented dialog understanding.
In this paper, we utilize semantic annotations in TOD and propose a new tree-induced semi-supervised contrastive learning paradigm to improve  dialog understanding.


\begin{table}[t]
\caption{Statistics of our pre-trained dialog dataset.}
\label{tab:data_stat}
\scalebox{0.9}{
\begin{tabular}{cccc}
\toprule
Statistics & \textit{AnPreDial} & \textit{UnPreDial} \\
\midrule
\# Datasets & 32 & 21  \\
\# Dialogs & 459,465 & 3,217,058 \\
\# Turns & 3,366,479 & 19,578,028 \\
Avg. tokens per turn & 13.9 & 14.5  \\
Avg. tokens per dialog & 101.8 & 88.2  \\
Total unique tokens & 46.8M & 283.7M \\
\bottomrule
\end{tabular}}
\end{table}

\begin{table*}[t]
\caption{Data Composition of \textit{AnPreDial} and \textit{UnPreDial}.}
\label{tab:data_comp}
\scalebox{0.8}{
\begin{tabular}{l|l}
\toprule
\textit{AnPreDial} & \begin{tabular}[c]{@{}l@{}}BANKING77 \cite{casanueva-etal-2020-efficient}, CLINC150 \cite{larson-etal-2019-evaluation}, HWU64 \cite{liu2021benchmarking}, REST8K \cite{coope-etal-2020-span}, TOP \cite{gupta-etal-2018-semantic-parsing}, ATIS \cite{hemphill-etal-1990-atis}, SNIPS \cite{coucke2018snips}, CrossNER \cite{liu2020crossner}, FB\_TOD\_SF \cite{schuster-etal-2019-cross-lingual},\\ MIT-restaurant \cite{liu2013asgard}, MIT-movies-eng \cite{liu2013asgard}, MIT-movies-trival10k13 \cite{liu2013asgard}, MultiWOZ\_coco \cite{li2020coco}, MultiWOZ \cite{eric-etal-2020-multiwoz}, STAR \cite{mosig2020star}, DailyDialog \cite{li2017dailydialog},\\ SGD \cite{rastogi2020towards}, Frames \cite{el-asri-etal-2017-frames}, MSRe2e \cite{li2018microsoft}, DSTC2 \cite{williams2016dialog}, DSTC3 \cite{williams2016dialog}, SimJoint \cite{shah-etal-2018-bootstrapping}, MulDoGo \cite{peskov-etal-2019-multi}, WOZ \cite{mrksic-etal-2017-neural}, TaskMaster1 \cite{byrne-etal-2019-taskmaster}, TaskMaster2 \cite{byrne-etal-2019-taskmaster},\\ TaskMaster3 \cite{byrne-etal-2019-taskmaster}, InCar \cite{eric-etal-2017-key}, MultiWOZ\_synthesis \cite{campagna-etal-2020-zero}, SwDA \cite{stolcke-etal-2000-dialogue}, BiTOD \cite{lin2021bitod}, PersuaGOOD \cite{wang-etal-2019-persuasion} \end{tabular} \\ 
\midrule
\textit{UnPreDial} & \begin{tabular}[c]{@{}l@{}}MulDoGo\_un \cite{peskov-etal-2019-multi}, ABCD \cite{chen2021action}, AirDialog \cite{wei-etal-2018-airdialogue}, CCPE \cite{radlinski-etal-2019-coached}, MetalWOZ \cite{shalyminov2020fast}, CMU\_DoG \cite{zhou-etal-2018-dataset}, ConvQuestions \cite{kacupaj-etal-2021-conversational}, CoQA \cite{reddy2019coqa}, CoSQL \cite{yu-etal-2019-cosql},\\ doc2dial \cite{feng-etal-2020-doc2dial}, DSTC10-track2 \cite{kim2021how}, DSTC10-track3 \cite{kottur2021simmc}, MedicalDialog \cite{zeng-etal-2020-meddialog}, Self-Dialog \cite{fainberg2018talking}, WOW \cite{dinan2018wizard}, TopicChat \cite{gopalakrishnan2019topical}, Persona-Chat \cite{zhang-etal-2018-personalizing},\\ MMD \cite{1704.00200}, CSQA \cite{1801.10314}, AmazonQA \cite{gupta2019amazonqa}, ChitChat \cite{will2020conversational} \end{tabular} \\ 
\bottomrule
\end{tabular}}
\end{table*}

\section{Pre-training Data Construction}
Previous TOD datasets are usually small and scattered since obtaining and labeling such data is time-consuming and labor-intensive.
In this paper, we collect new dialog datasets used for pre-training our SPACE-3 model, including a labeled dialog corpus (\textit{AnPreDial}) and a large-scale unlabeled dialog corpus (\textit{UnPreDial}).
The total statistics of our collected dialog corpora are shown in Table \ref{tab:data_stat}. More details about data composition can be found in Table \ref{tab:data_comp}.

\subsection{Labeled Dataset: \textit{AnPreDial}}
\label{sec:semantic_tree}
\begin{figure}[t]
    \centering
    \includegraphics[width=0.4\textwidth]{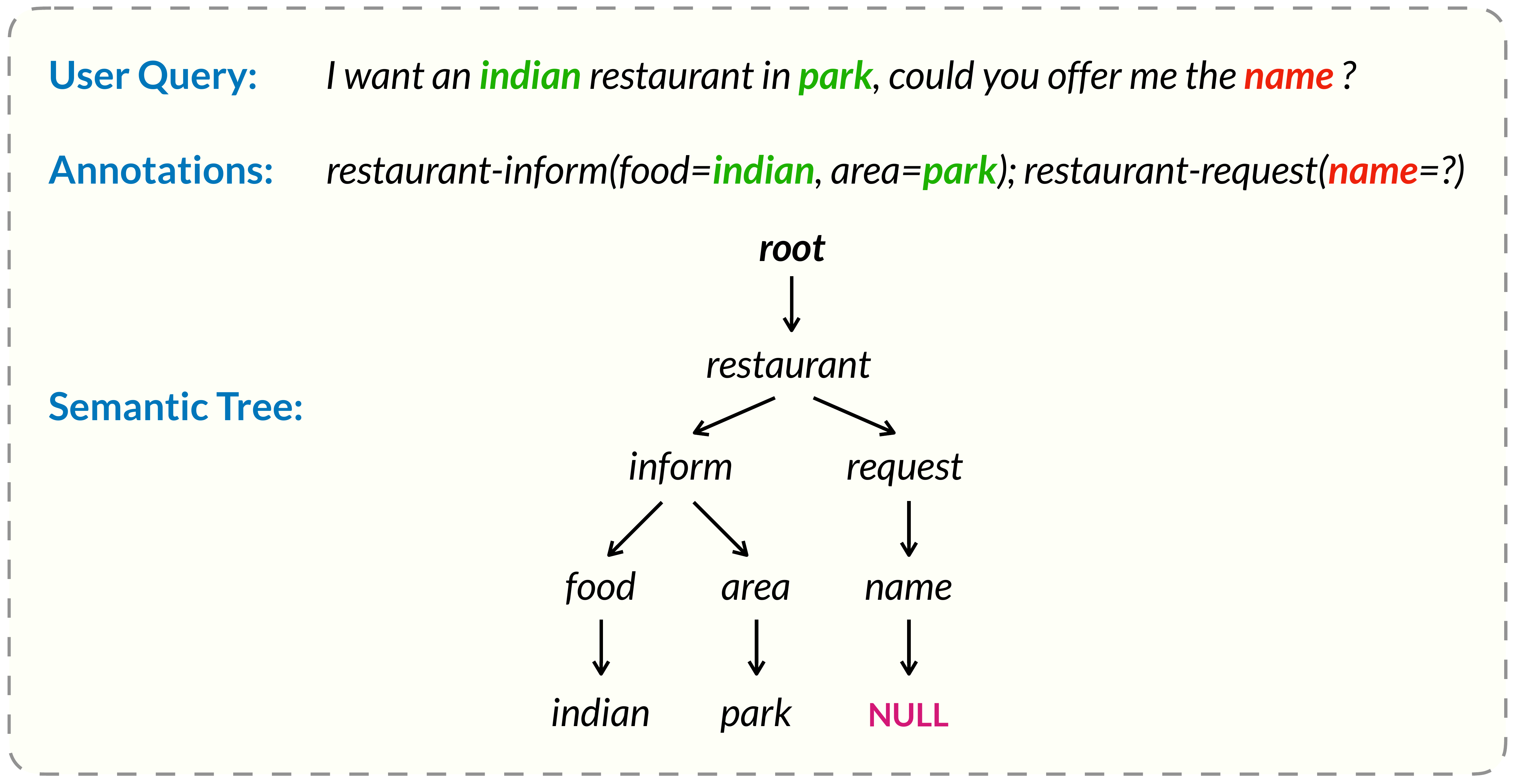}
    \caption{An example of user query with its semantic annotations in the labeled dataset \textit{AnPreDial}.}
    \label{fig:tree}
\end{figure}

As revealed by previous works, labeled data can not only accelerate the pre-training procedure but also improve the model performance \cite{assran2020supervision, khosla2020supervised, dai2021unimoco}.
Motivated by this observation, we exploit extra dialog annotations to learn better pre-trained dialog representations.
Specifically, task-oriented dialogs often involve several user goals, and the TOD systems need to help fulfill these goals through interaction with users. 
To this end, the user queries and system responses are usually composed of rich semantic annotations, including \texttt{DOMAIN}, \texttt{INTENT}, \texttt{SLOT} and \texttt{VALUE} labels.
For instance, given a user utterance \textit{``I want an indian restaurant in park, could you offer me the name?"}, the semantic annotations are \textit{``restaurant-inform(food=indian, area=park); restaurant-request(name=?)''}.
As shown in Figure \ref{fig:tree}, such semantic annotations are naturally structured, which could be expanded into tree structures called semantic trees.
A semantic tree typically contains four layers: \texttt{DOMAIN} layer, \texttt{INTENT} layer, \texttt{SLOT} layer, and \texttt{VALUE} layer.
Each layer regards respective annotation elements as nodes.
If no matched annotations are provided, the nodes of the corresponding layer are set as empty (denoted as \texttt{NULL}).
The first layer consists of \texttt{DOMAIN} nodes as successors of the \texttt{ROOT} node, preceded by \texttt{INTENT} nodes.
As children of \texttt{INTENT} nodes, \texttt{SLOT} nodes occupy the third layer of the semantic tree.
As leaf nodes, \texttt{VALUE} nodes take \texttt{SLOT} nodes as parents.

To provide sufficient high-quality dialog corpora with structured semantic annotations, we carefully examine currently available sources and aggregate them together into a large dataset called \textit{AnPreDial}, which contains 32 existing labeled TOD datasets with 3 million turns, ranging from single-turn QA to multi-turn dialogs.

\subsection{Unlabeled Dataset: \textit{UnPreDial}}
We aggregate 21 online dialog corpora and build a large-scale unlabeled dialog corpus (\textit{UnPreDial}) with 19 million utterances, ranging from online forums to conversational machine reading comprehension.
Due to the massive noises in \textit{UnPreDial}, we perform careful data cleaning, including:
(1) removing the dialogs when there is a URL in the utterances; 
(2) removing the dialogs that contain at least three repetitive words; 
(3) removing non-English sentences; 
(4) removing the sentences that contain special markers such as ``['' or ``]'', since these sentences could be markup;
(5) removing the dialogs that contain offensive language;
(6) replacing the non-unicode characters such as emojis.

\begin{figure*}[t]
    \centering
    \includegraphics[width=0.7\textwidth]{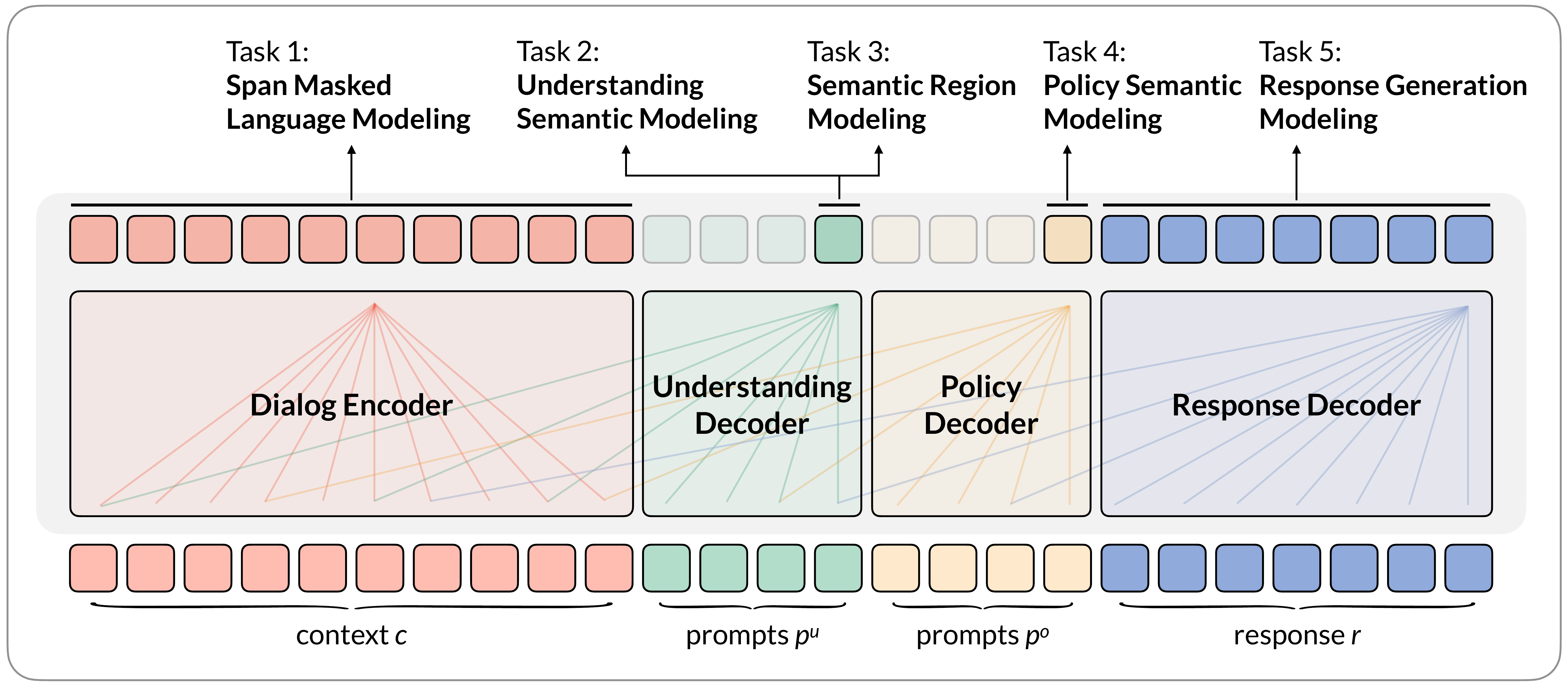}
    \caption{
    The architecture of proposed SPACE-3 model, which successively includes a dialog encoder, a understanding decoder, a policy decoder and a response decoder.
    In each module, the lines with different colors denote the self-attention positions where current token can attend.
    The encoder is bi-directional, while three decoders are uni-directional.
    As extra inputs of the understanding and policy decoders, two kinds of prompts are leveraged to extract semantics and help pass the task-flow in a TOD system.
    Finally, five pre-training objectives are carried out to optimize SPACE-3 jointly in a multi-task paradigm.
    }
    \label{fig:model}
\end{figure*}

\begin{figure}
    \centering
    \includegraphics[width=0.35\textwidth]{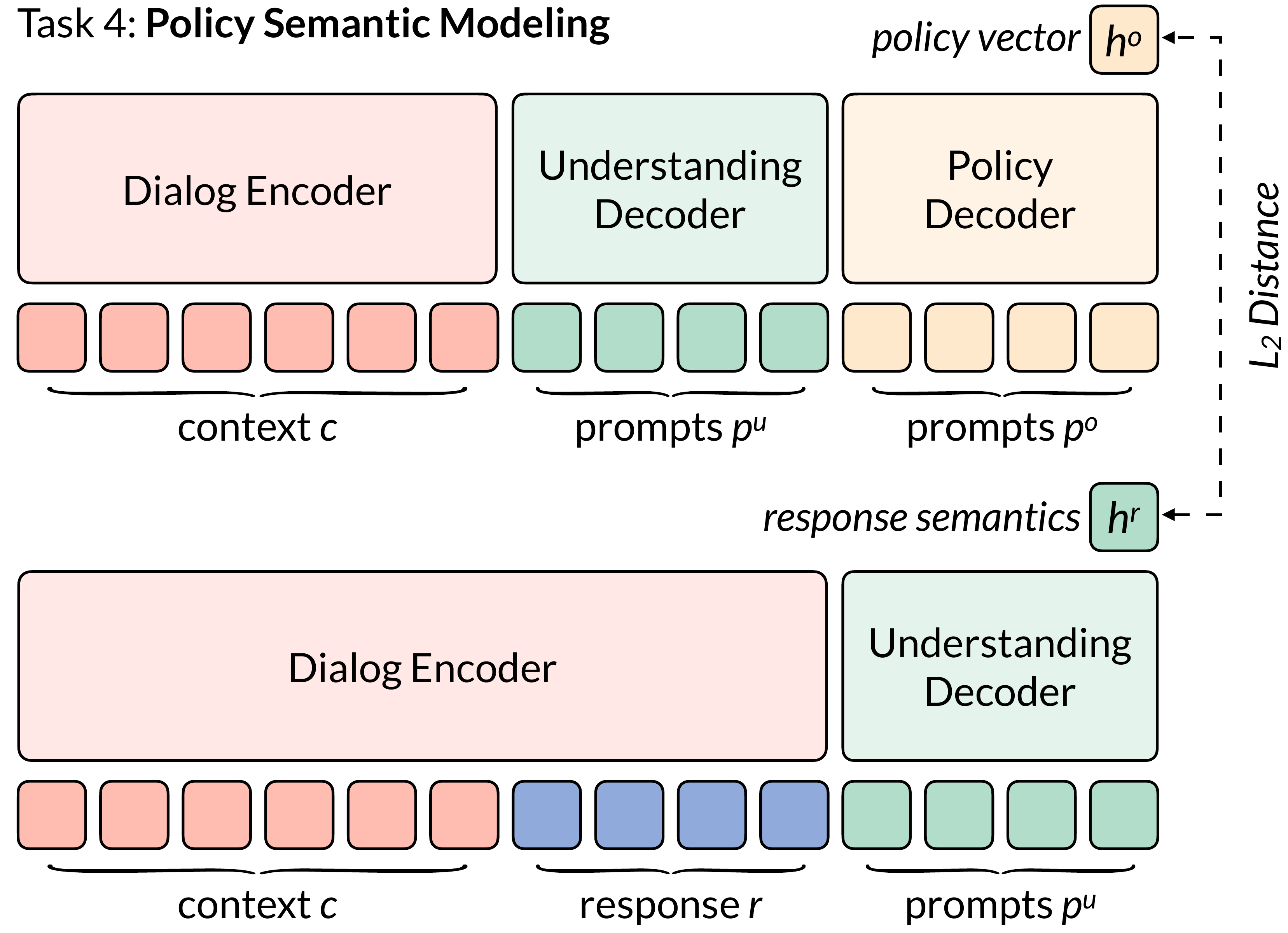}
    \caption{
    The procedure of policy semantic modeling, minimizing the $\mathcal{L}_2$ distance between the predicted policy vector $h^o$ and the extracted response semantic vector $h^r$.
    }
    \label{fig:policy}
\end{figure}

\section{Method}
In this section, we first introduce the model architecture.
Then we elaborate on each objective used in our pre-training.

\subsection{Model Architecture}
We choose UniLM \cite{dong2019unified} as our backbone model for task-oriented dialog modeling.
UniLM contains a bi-directional Transformer encoder and a uni-directional Transformer decoder with shared parameters, which is naturally suitable for both dialog understanding and dialog generation.
As illustrated in Figure \ref{fig:model}, our proposed SPACE-3 architecture consists of four successive components to maintain the task-flow in TOD systems, including a dialog context encoder, a dialog understanding decoder, a dialog policy decoder and a dialog generation decoder.
Similar to UniLM, we share parameters of all modules and adopt different self-attention mechanisms to distinguish the encoder and decoders.

Let $D=\{d_i=<u_i, s_i>\}^{K}_{i=1}$ denotes a conversation with $K$ dialog turns.
$u_i$ and $s_i$ denote the user utterances and system utterances, respectively.
At current turn $k$, we define the dialog context $c$ as the concatenation of utterances $\{d_j\}^{k-1}_{j=1}$ and user query $u_k$.
To simplify notations, we use $q=\{q_1, q_2, ..., q_{M}\}$ and $r=\{r_1, r_2, ..., r_{N}\}$ to represent the user query $u_k$ and system response $s_k$, where $q_m$ denotes the $m$-th token in $u_k$ and $r_n$ denotes the $n$-th token in $s_k$. $M$ and $N$ denote corresponding utterance lengths.
Next, we describe the four modules and the input representations in detail.

\subsubsection{Input Representation}
Following previous work \cite{bao-etal-2020-plato}, for each input token, its vector representation is computed by summing the corresponding token embedding, the role embedding, the turn embedding and the position embedding. 
The role embeddings are used to differentiate the characters evolved in the current utterance (i.e., $E_U$ for user role while $E_S$ for system role).
The turn embeddings are assigned to each token according to its relative turn order in multi-turn utterances.
The position embeddings are assigned to each token according to its relative position within the corresponding utterance.
In addition, we add special tokens \texttt{[BOU]}, \texttt{[EOU]} to bound the user utterance, and \texttt{[BOS]}, \texttt{[EOS]} to bound the system utterance.
It is noteworthy that the role, turn and position embeddings are set to empty for the prompt tokens. 

\subsubsection{Dialog Context Encoder}
To overcome the challenge of modeling long dialog history in multi-turn conversations, the dialog context encoder employs a bi-directional Transformer to encode the context utterances $c$, which can capture the common and essential information for subsequent modules.

\subsubsection{Dialog Understanding Decoder}
\label{sec:representation}
We leverage a dialog understanding decoder (a uni-directional Transformer) to extract semantic representations of current user query $q$ conditioned on the dialog context $c$, which are used for classification tasks such as intent recognition.
Therefore, it is vital to map multiple utterances to a semantically meaningful representation.
Prior work \cite{clark-etal-2019-bert} revealed that transformer-based models tend to attribute a significant amount of attention to the \texttt{[CLS]} token, resulting in diluted representations.
Instead of using the latent representation of the \texttt{[CLS]} token, \cite{mehri-eric-2021-example} proposed to have observer tokens that attend to the input words but are not being attended to by the model. 
Inspired by the observers \cite{mehri-eric-2021-example} and prompt tuning \cite{lester-etal-2021-power}, we take the concatenation of a sequence of soft prompt tokens and the dialog context $c$ as input of the dialog understanding decoder to extract the rich semantic representations of dialog context.
To be specific, we use $p^u=\{p^u_{1}, p^u_2, ..., p^u_{A}\}$ to represent the dialog understanding prompt sequence, where $A$ is the length of the prompt.
Different from \cite{mehri-eric-2021-example} that directly average observers' final representations from the last transformer layer, we select the hidden state of the last prompt token $p^u_{A}$ as the sentence representation $h^q$ of the user query $q$ due to the uni-directional sequential property.
As shown in Figure \ref{fig:policy}, conditioned on the concatenation of $c$, $r$ and $p^u$, we can also obtain the hidden state of the prompt token $p^u_{A}$ as the semantic representation $h^r$ of the response $r$ for better understanding the system behaviors.

\subsubsection{Dialog Policy Decoder}
The dialog policy decoder (a uni-directional Transformer) is devised to determine high-level behaviors of the system, which leads to successful task completion in TOD systems.
Similar to the dialog understanding prompts, we also adopt a policy prompt sequence, which is denoted as $p^o=\{p^o_{1}, p^o_2, ..., p^o_{B}\}$ where $B$ denotes the length of the policy prompt sequence. 
Taking $(c, p^u, p^o)$ as input, the dialog policy decoder obtains the hidden state of the prompt token $p^o_{B}$ as the output policy vector $h^o$.

\subsubsection{Dialog Generation Decoder}
Given the dialog context $c$, the understanding prompt sequence $p^u$ and policy prompt sequence $p^o$, we apply the dialog generation decoder to generate an appropriate response $r$ in an auto-regressive manner, returning the reply to the user at the current dialog turn.

\subsection{Pre-training Objectives}
We employ five objectives to pre-train SPACE-3 in a multi-task learning paradigm, including span masked language modeling, semi-supervised understanding semantic modeling, semantic region modeling, policy semantic modeling and response generation modeling.

\subsubsection{Span Masked Language Modeling}
Previous work \cite{devlin-etal-2019-bert} show that masked language modeling (MLM) is an effective pre-training strategy for distributed representation learning, in which a random sample of tokens in the input sequence is replaced with the special token \texttt{[MASK]} and the model needs to predict the masked tokens.
Span Masking \citep{joshi-etal-2020-spanbert} improves the original MLM by masking random contiguous spans rather than random individual tokens.
In this paper, we employ the span masking language modeling (SLM) to pre-train the dialog context encoder on both the labeled dataset \textit{AnPreDial} and the unlabeled dataset \textit{UnPreDial}. Specifically, for labeled data annotated with slot-value pairs, we mask all value spans in $c$ to construct the perturbed input sequence; for unlabeled data, we randomly mask contiguous text spans instead. Then the cross-entropy loss is employed to minimize the negative log-likelihood of the masked tokens, which is defined as:
\begin{equation}
    \mathcal{L}_{\mathrm{slm}}=-\sum_{\hat{c} \in m\left(c\right)} \log p\left(\hat{c} \mid c_{\backslash m\left(c\right)}\right)
\end{equation}
where $m(c)$ denotes the masked tokens from $c$ and ${c}_{\backslash m(\mathbf{c})}$ indicates the rest tokens except $m(c)$.

\subsubsection{Semi-supervised Understanding Semantic Modeling}
\label{sec:understanding}
To enhance dialog representation learning with structured semantic annotations, we perform semi-supervised pre-training on both the labeled data and unlabeled data. 
Specifically, we adopt a novel tree-induced supervised contrastive objective on the labeled dataset \textit{AnPreDial}, while a vanilla self-supervised contrastive objective is applied on the unlabeled dataset \textit{UnPreDial}. 
For the labeled data, we treat all data in the same batch as positive samples, which are assigned with different similarity scores.
As depicted in Section \ref{sec:semantic_tree}, we reorganize the structured semantic annotations stored in labeled dialog corpora as semantic trees.
According to the semantic trees' similarity, we estimate the similarity of corresponding dialog samples.
For each pair of semantic trees, we leverage the Tree Editing Distance to calculate their similarity score, which is considered to be the weighted number of edit operations (insert, delete, and modify) to transform one tree to another.
Different from \cite{zhang1989simple} that computes the distance between two ordered trees, we do not differentiate the order of the children of a node.
For example in Figure \ref{fig:tree}, as children of the \textit{inform} node, the left-to-right arrangement of \textit{<food, area>} is equivalent to the arrangement of \textit{<area, food>}.
Therefore, we reorder all children nodes recursively according to the  alphabetical order of their annotation strings.
Given a pair of reordered semantic trees (denoted as $T_i,T_j$),  we calculate the similarity coefficient $f_{i, j}$ by:
\begin{gather}
\label{eq:norm}
    f_{i, j} = \frac{\max\left\{|T_i|, |T_j|\right\}-d_{i, j}}{\max\left\{|T_i|, |T_j|\right\}}
    \\
    d_{i, j} = \mathrm{TreeEditingDistance}\left(T_i, T_j\right)
\end{gather}
where $\mathrm{TreeEditingDistance}$ is the function to calculate the tree editing distance and $|\cdot|$ denotes the number of nodes in the semantic tree. The similarity coefficient $f_{i, j} \in \left[0, 1\right]$ as the distance $d_{i, j} \in \left[0, \max\left\{|T_i|,|T_j|\right\}\right]$.

As mentioned in Section \ref{sec:representation}, we use the hidden states of the last tokens in the respective prompt sequences to represent the sentence representations $h^q, h^r \in \mathbb{R}^{768}$ of user query and system response, respectively. 
Thus for each pair of dialog samples $(i, j)$, their output sentence representations are $h^q_i, h^r_i$ for the sample $i$ and $h^q_j, h^r_j$ for the sample $j$. Here, we use the superscript $^q$ and $^r$ to represent the variables related to the user query and the system response, respectively.  
Hence, we use $f^q_{i,j}$ and $f^r_{i,j}$ to denote the similarity scores between semantic trees of user queries and system responses, respectively. 
Note that we use the same way to compute the semi-supervised pre-training objective for user queries or system responses.
For simplicity, we only introduce the pre-training objective on the user side as an example.
Suppose the size of the current batch is $L$. We follow the common practice in \cite{gao2021simcse} to duplicate the training data via  dropout-based data augmentation. 
In this way, we acquire a new $2L$-length batch. 
Let $I=\{1\dots2L\}$ be the index set of the new batch, and we adopt the tree-induced supervised contrastive objective for labeled data as follows:
\begin{equation}
\small
\mathcal{L}^q_\mathrm{sup} = - \sum_{i \in I} \sum_{j \in I}\frac{f^q_{i, j}}{\sum_{v \in I} f^q_{i, v}}  \log \frac{\exp \left(\sigma\left({h^q_i}\right)\cdot \sigma\left({h^q_j}\right) / \tau\right)}{\sum_{l \in I,l \neq i} \exp \left(\sigma\left({h^q_i}\right) \cdot \sigma\left({h^q_l}\right)/ \tau\right)}
\end{equation}
where $\sigma$ is a transformation function which is defined by:
\begin{equation}
\sigma\left(h^q\right) = \mathrm{Norm}\left(W_1 h^q + b_1\right)
\end{equation}
where  $W_1\in\mathbb{R}^{768 \times 768}$ and $b_1\in\mathbb{R}^{768 \times 1}$ are learnable parameters. 
$\mathrm{Norm(\cdot)}$ is the normalization operation to constrain $||\mathrm{Norm(\cdot)||}=1$ and $\tau\in[0,1]$ is a temperature hyper-parameter. 
Note that our method is quite different from the common supervised contrastive learning \cite{vulic2021convfit} in which only augmented data or data from the exact same label are considered as positive samples, and all other data are considered as negative samples.
Instead, we regarded all in-batch data as positive samples, assigned with different similarity scores ranging from $0$ to $1$.

For unlabeled data, we do not use semantic trees due to the lack of available semantic annotations.
Instead, we adopt a self-supervised contrastive objective, where only the augmented data by dropout is treated as a positive sample \cite{gao2021simcse}.
The loss function is defined as:
\begin{equation}
\small
\mathcal{L}^q_\mathrm{self} = -\sum_{i \in I} \sum_{j \in I} \textbf{1}\left(j=i^+\right) \log \frac{\exp \left(\sigma\left({h^q_i}\right)\cdot \sigma\left({h^q_j}\right) / \tau\right)}{\sum_{l \in I,l \neq i} \exp \left(\sigma\left({h^q_i}\right) \cdot \sigma\left({h^q_l}\right)/ \tau\right)}
\end{equation}
where $\textbf{1}(j=i^+)$ indicates that it takes 1 if and only if the data sample $j$ is the augmentation ($i^+$) of the data sample $i$; otherwise 0 is obtained.

Overall, the tree-induced semi-supervised contrastive learning objective for the understanding of the user query is:
\begin{equation}
    \mathcal{L}^q_\mathrm{scl} = \mathcal{L}^q_\mathrm{sup} + \mathcal{L}^q_\mathrm{self}
\end{equation}
Combined with the objective $\mathcal{L}^r_\mathrm{scl}$ on the system side which is calculated in the same way as $\mathcal{L}^q_\mathrm{scl}$, our total loss for the dialog understanding modeling is computed as:
\begin{equation}
    \mathcal{L}_\mathrm{scl} = \mathcal{L}^q_\mathrm{scl} + \mathcal{L}^r_\mathrm{scl}
\end{equation}

\subsubsection{Semantic Region Modeling}
\label{sec:bow}
We additionally design a bag-of-words loss to eliminate the influence of uninformative utterances in the long dialog context $c$ using the prompts $p^u$.
This objective is to optimize the log-likelihood of $q_{\rm bow}$ which is the variant of $q$ without considering word order:
\begin{gather}
    \mathcal{L}^q_\mathrm{bow} 
    = -\log p\left(q_{\rm bow} \mid c, p^u \right)  
    = -\log \prod_{m=1}^{M} H^{'}_{\textit{index}\left(q_m\right)}
    \\
    H^{'} = \operatorname{softmax}\left(W_2 h^q + b_2\right) \in \mathbb{R}^{|V|}
\end{gather}
where $|V|$ refers to the vocabulary size and $\textit{index}\left(q_m\right)$ is the word index in the vocabulary of the $m$-th word $q_m$ in the utterance $q$.
$H^{'}_{\textit{index}\left(q_m\right)}$ denotes the estimated probability of $q_m$.
$W_2$ and $b_2$ are learnable parameters.
The bag-of-words loss discards the order of words and predicts the words in $q$ in a non-autoregressive way, which could force the semantic representation to capture the global information of the given utterance.

In addition, we also compute the bag-of-words loss on the top of the representation $h^r$ for the system response $r$, which is to predict the unordered words $r_{\rm bow}$ as follows:
\begin{gather}
    \mathcal{L}^r_\mathrm{bow}
    = -\log p\left(r_{\rm bow} \mid c, r, p^u \right) 
    = -\log \prod_{n=1}^{N} H^{''}_{\textit{index}\left(r_n\right)}
    \\
    H^{''} = \operatorname{softmax}\left(W_2 h^r + b_2\right) \in \mathbb{R}^{|V|}
\end{gather}
Totally, the loss for semantic region modeling is defined as:
\begin{equation}
    \mathcal{L}_\mathrm{bow} = \mathcal{L}^q_\mathrm{bow} + \mathcal{L}^r_\mathrm{bow}
\end{equation}

\subsubsection{Policy Semantic Modeling}
The dialog policy module determines high-level dialog acts of TOD systems towards successful task completion. SPACE-3 predicts the policy vector $h^o$ to determine the high-level semantics of the response $r$ conditioned on the context $c$, the dialog understanding prompts $p^u$,  and the policy prompts $p^o$.
It is intuitive that the meanings of the learned policy vector should be close to the meanings of semantic vectors inferred from the response as a posterior. To this end, we first extract the semantic response vector $h^r$ based on the sequence ($c$, $r$, $p^u$) and the policy vector $h^o$  based on the the sequence ($c$, $p^u$, $p^o$). Then we decrease their discrepancy by minimizing the $L_2$ distance.
As illustrated in Figure \ref{fig:policy}, we define the policy semantic modeling objective $\mathcal{L}_{\rm psm}$ as follows:
\begin{equation}
    \mathcal{L}_\mathrm{psm} = ||h^o-h^r||^2_2
\end{equation}

\subsubsection{Response Generation Modeling}
Based on the dialog context $c$, understanding prompts  $p^u$ and policy prompts $p^o$, the response generation task is carried out to produce the appropriate system response $r$ in an auto-regressive manner.
We adopt the standard negative log-likelihood loss $\mathcal{L}_\mathrm{rgm}$ for the generation task:
\begin{equation}
\mathcal{L}_\mathrm{rgm} =-\sum_{n=1}^{N} \log p\left(r_n \mid c, p^u, p^o, r_{<n}\right)
\end{equation}
where $r_n$ is the $n$-th word in $r$ and $r_{<n}=\{r_1, ..., r_{n-1}\}$ represents the words of previous steps.

\subsubsection{Joint Training}
Our unified pre-training strategy allows the TOD systems to learn the dialog understanding, policy planning, and dialog generation abilities from large-scale dialog corpora via multi-task learning.
To sum up, the overall training objective $\mathcal{L}_\mathrm{joint}$ of SPACE-3 can be defined as follows:
\begin{equation}
    \mathcal{L}_{\rm joint} = 
    \mathcal{L}_\mathrm{slm} + 
    \mathcal{L}_\mathrm{scl} + 
    \mathcal{L}_\mathrm{bow} + 
    \mathcal{L}_\mathrm{psm} + 
    \mathcal{L}_\mathrm{rgm}
\end{equation}

\subsection{Fine-tuning}
In the fine-tuning stage, we concentrate on task-oriented dialog tasks, including dialog understanding, policy optimization, and dialog generation. 
For dialog understanding tasks, we only leverage the dialog encoding module and the dialog understanding module to fine-tune our pre-trained model, discarding the dialog policy and generation modules.
Thus only the understanding prompt sequence $p^u$ is employed to extract sentence representations, which are used for classification tasks such as intent recognition or dialog state tracking.
For policy optimization and dialog generation tasks, we retain the integrated model architecture to produce appropriate responses, where two kinds of prompts $p^u$ and $p^o$ are adopted to characterize the dialog understanding and dialog policy for better dialog generation.

\section{Experimental Setup}
In this section, we will introduce experimental data, compared baselines and the implementation details. 

\subsection{Experimental Data}
To comprehensively evaluate our SPACE-3, we carefully choose \textbf{eight} downstream tasks in terms of dialog understanding, policy optimization and response generation. For the understanding modeling, we have three intent prediction datasets and one dialog state tracking dataset. For policy and generation modeling, we follow the setup in previous work \cite{lubis-etal-2020-lava} and evaluate the proposed model on four well-studied end-to-end dialog modeling datasets. 

\paragraph{\textbf{Intent Prediction (IP)}} 
The goal of IP is to predict which intention the user expresses in the current utterance, which is a single-turn classification problem. Recently, three large benchmarks have been used for evaluating IP models, including (i) \textit{BANKING77}  \cite{casanueva-etal-2020-efficient}, a single turn dataset in the banking domain with
77 intent labels and 13,083 utterances; (ii) \textit{CLINIC150} \cite{larson-etal-2019-evaluation}, a multi-domain dataset that contains
23,700 utterances, 150 different intent classes and 10 domains; (iii) \textit{HWU64} \cite{liu2021benchmarking} includes 25,716 utterances for 64 intents spanning 21 domains. To conduct the downstream task,  a  linear layer neural network is added to the pooled representation output $z$ for intent prediction. Following previous work \cite{mehri2020dialoglue}, we set the evaluation metric as turn accuracy (\texttt{ACC}). 

\paragraph{\textbf{Dialog State Tracking (DST)}}
DST is a multi-turn slot-filling task, where the model needs to maintain a dialog state in terms of slot-value pairs throughout the duration of a dialog. MultiWOZ \cite{budzianowski-etal-2018-multiwoz} is one of the most challenging datasets for DST, which contains 10k dialogs spanning 7 distinct domains. In particular we choose the \textit{MultiWOZ2.2} version \cite{zang-etal-2020-multiwoz} as our benchmark since it has fixed a large proportion of dialog state annotation errors. Following \cite{heck-etal-2020-trippy, mehri2020dialoglue, dai-etal-2021-preview}, we adopt the joint goal accuracy (\texttt{JGA}) as our evaluation metric, counting the ratio of turns that model gives correct predictions for all slots. For a fair comparison, all compared pre-trained conversation models are used to replace the original BERT part in the TripPy model \cite{heck-etal-2020-trippy} for downstream fine-tuning in our experiments.

\paragraph{\textbf{End-to-End Dialog Modeling (E2E)}}
This is a standard task to evaluate the capability of both policy optimization and dialog generation. To be specific, we evaluate the proposed model on four popular benchmark task-oriented dialog datasets: the Cambridge Restaurant (\textit{CamRest676}) \cite{wen-etal-2017-network}, Stanford In-Car Assistant (\textit{In-Car}) \cite{eric-etal-2017-key}, \textit{MultiWOZ2.0} and \textit{MultiWOZ2.1}. CamRest676 contains single-domain dialogs belonging to the restaurant domain. In-Car consists of dialogs between a user and a in-car assistant system covering several tasks, including calendar scheduling, weather information retrieval and point-of-interest navigation. MultiWOZ2.1 is a newer version of MultiWOZ2.0 where the wrong dialog actions have been rectified. Both datasets are widely used for end-to-end dialog evaluation.
We use \texttt{BLEU} \cite{papineni-etal-2002-bleu} to measure the response generation quality. For MultiWOZ, we report the metrics of whether the system provides an correct entity (\texttt{Inform}) and  answers all the requested information (\texttt{Success}). A combined score (\texttt{Comb}) \cite{mehri-etal-2019-structured} is computed by (\texttt{Inform+Success})$\times$0.5+\texttt{BLEU} as an overall quality measure. Similarly, for In-Car and CamRest676, we use \texttt{Match} and \texttt{SuccF1} following \cite{lei-etal-2018-sequicity}, and calculate a  combined score (\texttt{Comb}) via (\texttt{Match+SuccF1})$\times$0.5+\texttt{BLEU}.



\begin{table*}[t]
\caption{Comparison of different pre-trained conversation models on eight dialog benchmarks under full-data and few-shot settings. $*$ denotes continually pre-traning on our pre-trained dialog corpora. $\dagger$ denotes results from the original paper. We calculate \texttt{ACC}, \texttt{JGA} and \texttt{Comb} metrics for IP, DST and E2E tasks respectively and average them for the overall scores.}
\label{tab:overall}
\scalebox{0.9}{
\begin{tabular}{cccccccccc}
\toprule
\multicolumn{1}{c|}{\multirow{2}{*}{Model}} & \multicolumn{3}{c|}{IP} & \multicolumn{1}{c|}{DST} & \multicolumn{4}{c|}{E2E} & \multirow{2}{*}{Average} \\ \cline{2-9}
\multicolumn{1}{c|}{} & BANKING77 & CLINIC150 & \multicolumn{1}{c|}{HWU64} & \multicolumn{1}{c|}{MultiWOZ2.2} & CamRest & In-Car & MultiWOZ2.0 & \multicolumn{1}{c|}{MultiWOZ2.1} &  \\ \hline
\multicolumn{1}{c|}{TOD-BERT} & 92.76 & 96.64 & \multicolumn{1}{c|}{91.91} & \multicolumn{1}{c|}{54.26} & 113.28 & 103.64 & 100.88 & \multicolumn{1}{c|}{101.23} & 94.33 \\
\multicolumn{1}{c|}{PLATO} & 92.56 & 96.00 & \multicolumn{1}{c|}{89.96} & \multicolumn{1}{c|}{53.89} & 114.25 & 104.07 & 102.16 & \multicolumn{1}{c|}{102.98} & 94.48 \\
\multicolumn{1}{c|}{PPTOD} & 93.86$\dagger$ & 97.13 & \multicolumn{1}{c|}{92.84} & \multicolumn{1}{c|}{55.26} & 114.92 & 104.68 & 102.92$\dagger$ & \multicolumn{1}{c|}{102.26$\dagger$} & 95.48 \\
\multicolumn{1}{c|}{UniLM$^*$} & 93.12 & 96.11 & \multicolumn{1}{c|}{90.15} & \multicolumn{1}{c|}{54.25} & 115.03 & 104.91 & 105.35 & \multicolumn{1}{c|}{105.94} & 95.61 \\
\multicolumn{1}{c|}{PPTOD$^*$} & 93.64 & 97.29 & \multicolumn{1}{c|}{92.94} & \multicolumn{1}{c|}{56.01} & 115.73 & 106.01 & 107.14 & \multicolumn{1}{c|}{107.22} & 97.00 \\
\multicolumn{1}{c|}{SPACE-3} & \textbf{94.94} & \textbf{97.89} & \multicolumn{1}{c|}{\textbf{94.14}} & \multicolumn{1}{c|}{\textbf{57.50}} & \textbf{116.67} & \textbf{107.13} & \textbf{110.95} & \multicolumn{1}{c|}{\textbf{110.76}} & \textbf{98.75} \\ \hline
\multicolumn{10}{c}{Few-shot setting (10\%)} \\ \hline
\multicolumn{1}{c|}{TOD-BERT} & 85.19 & 94.18 & \multicolumn{1}{c|}{87.36} & \multicolumn{1}{c|}{47.67} & 101.89 & 92.18 & 85.17 & \multicolumn{1}{c|}{86.03} & 84.96 \\
\multicolumn{1}{c|}{PLATO} & 84.03 & 93.33 & \multicolumn{1}{c|}{86.71} & \multicolumn{1}{c|}{46.98} & 102.98 & 93.14 & 90.72 & \multicolumn{1}{c|}{90.31} & 86.03 \\
\multicolumn{1}{c|}{PPTOD} & 82.81$\dagger$ & 94.27 & \multicolumn{1}{c|}{87.36} & \multicolumn{1}{c|}{46.98} & 103.88 & 93.89 & 91.96$\dagger$ & \multicolumn{1}{c|}{92.05} & 86.65 \\
\multicolumn{1}{c|}{UniLM$^*$} & 85.32 & 94.00 & \multicolumn{1}{c|}{87.08} & \multicolumn{1}{c|}{46.88} & 103.42 & 92.97 & 90.96 & \multicolumn{1}{c|}{90.20} & 86.35 \\
\multicolumn{1}{c|}{PPTOD$^*$} & 87.63 & 94.56 & \multicolumn{1}{c|}{88.57} & \multicolumn{1}{c|}{48.24} & 105.01 & 94.01 & 93.80 & \multicolumn{1}{c|}{94.22} & 88.26 \\
\multicolumn{1}{c|}{SPACE-3} & \textbf{88.34} & \textbf{95.00} & \multicolumn{1}{c|}{\textbf{89.22}} & \multicolumn{1}{c|}{\textbf{50.66}} & \textbf{106.79} & \textbf{98.30} & \textbf{98.88} & \multicolumn{1}{c|}{\textbf{98.74}} & \textbf{90.74} \\ \bottomrule
\end{tabular}}
\end{table*}

\subsection{Baselines}
All compared baselines can be divided into two categories: (i) the pre-trained conversation models that are adapted for the eight downstream dialog tasks; and (ii) the specific models that are tailored to certain tasks such as intent prediction.

\paragraph{\textbf{Baselines for Overall Comparison}}
We compare SPACE-3 with three strong pre-trained conversation models: 1) \textbf{TOD-BERT} \cite{wu-etal-2020-tod}, a BERT$_{\text{base}}$ model that has been continually pre-trained on large scale TOD datasets with MLM and response selection losses; 
2) \textbf{PLATO} \cite{bao-etal-2020-plato}, a pre-trained dialog generation model with discrete latent variable based on UniLM \cite{dong2019unified};
3) \textbf{PPTOD} \cite{su2021multitask}, a T5 \cite{JMLR:v21:20-074} model that has been continually pre-trained on eleven heterogeneous annotated TOD corpora under the same data format.

\paragraph{\textbf{Baselines for Specific Tasks}}
For intent prediction, we compare SPACE-3 with four state-of-the-art baselines: \textbf{ConvBERT} \cite{mehri2020dialoglue}, \textbf{Example+Observer} \cite{mehri-eric-2021-example}, \textbf{USE+ConveRT} \cite{casanueva-etal-2020-efficient}, and \textbf{ConvFit} \cite{vulic2021convfit}. 
For the dialog state tracking task, we compare with \textbf{DS-DST} \cite{zhang2019find}, \textbf{Seq2Seq-DU} \cite{feng-etal-2021-sequence}, \textbf{PLATO-XL} \cite{bao2021platoxl} and \textbf{AG-DST} \cite{tian-etal-2021-amendable}. 
For the end-to-end dialog modeling task, we have different baselines on different datasets for comparison. For MultiWOZ2.0 and MultiWOZ2.1, we compare with \textbf{SimpleTOD} \cite{hosseini2020simple}, \textbf{DoTS} \cite{jeon2021domain}, \textbf{UBAR}  \cite{yang2020ubar}, and \textbf{MTTOD}  \cite{lee-2021-mttod}. 
For CamRest676 and In-Car, we compare with \textbf{SEDST} \cite{jin2018explicit}, \textbf{TSCP} \cite{lei-etal-2018-sequicity}, \textbf{FSDM} \cite{shu2019flexibly} and \textbf{LABES} \cite{zhang-etal-2020-probabilistic}.


\subsection{Implementation Details}
In our experiments, the number of transformer blocks in SPACE-3 is 12 and the hidden embedding dimension is 768.
In the pre-training stage, SPACE-3 is initialized with UniLM.
The maximum sequence length of dialog context and response is set to 256 and 50. 
The sequence length of understanding prompts and policy prompts is set to 5.
The batch size is set to 128 and an AdamW optimizer is employed for optimization with an initial learning rate of 1e-5. 
The dropout rate is set to 0.2. 
In the fine-tuning stage, the grid search algorithm is applied on the validation set to tune hyper-parameters automatically.
For all downstream tasks, we do experiments with 3 random seeds and take the average as final results.

\section{Experimental Results}
\subsection{Overall Comparison with PCMs}
We first compare our SPACE-3 with three pre-trained conversation models (PCMs): TOD-BERT, PLATO and PPTOD. Since TOD-BERT is not designed for dialog generation, we adapt it into the same UniLM architecture with its own parameter initialization. We also compare with a UniLM to see the basic performance of the unified  encoder-decoder structure.
For fair comparison, we continually pre-trained the PPTOD and the UniLM on our pre-trained dialog corpora (\textit{AnPreDial} and \textit{UnPreDial}) to mitigate the data discrepancy, which are denoted with $^*$ in the Table \ref{tab:overall}. 
To extensively evaluate our model, we aggregate the results on eight different benchmarks, including intent prediction (IP), dialog state tracking (DST) and end-to-end dialog modeling (E2E) tasks. We calculate the turn accuracy (\texttt{ACC}), joint goal accuracy (\texttt{JGA}) and combined scores (\texttt{Comb}) for these tasks, and  compute the average score as the overall metric to measure the model performance. 
Due to few-shot learning gains increasing attention in various dialog tasks to assess model capability \citep{geng-etal-2019-induction, dai-etal-2020-learning, tseng-etal-2021-transferable}, we conduct experiments under both a full-data setting that uses all training data to fine-tune the models and a few-shot setting that uses only 10\% of training data to fine-tune the models. To avoid the data leakage in the few-shot setting, we only use the 10\% of the training data of the target evaluation dataset for pre-training if it exists in the \textit{AnPreDial}. 
As shown in Table \ref{tab:overall}, SPACE-3 outperforms all baselines on all datasets. On the full-data setting and the few-shot setting, SPACE-3 surpasses the PPTOD$^*$ with 1.75 and 2.48 absolute average score improvement respectively, indicating the SPACE-3 have better adaptability on all types of dialog tasks. The good results of UniLM$^*$ show that its basic model architecture is well suitable for both dialog understanding and generation, but SPACE-3 achieves much better performance due to our pre-trained objectives can learn better characteristics of task-flow in TOD. 

\begin{table}[t]
    \caption{Comparison results on dialog understanding  tasks.}
    \label{tab:intent}
   \scalebox{0.9}{
    \begin{tabular}{c|ccc}
    \toprule
    Model & BANKING77 & CLINIC150 & HWU64 \\ \hline
    ConvBERT & 92.95 & 97.07 & 90.43 \\
    USE+ConvRT & 93.36 & 97.16 & 92.62 \\
    Example+Observer & 93.83 & 97.31 & 93.03 \\
    ConvFiT & 94.16 & 97.34 & 92.42 \\
    SPACE-3 & \textbf{94.94} & \textbf{97.89} & \textbf{94.14} \\ \hline\hline
    Model & \multicolumn{3}{c}{MultiWOZ2.2} \\ \hline
    DS-DST & \multicolumn{3}{c}{51.7} \\
    Seq2Seq-DU & \multicolumn{3}{c}{54.40} \\
    PLATO-XL & \multicolumn{3}{c}{57.16} \\
    AG-DST & \multicolumn{3}{c}{57.26} \\
    SPACE-3 & \multicolumn{3}{c}{\textbf{57.50}} \\ \bottomrule
    \end{tabular}}
\end{table}

\subsection{Model Comparison on Individual Tasks}
We also compare SPACE-3 with state-of-the-art methods on individual tasks. Generally, these strong baselines are tailored to a specific task and cannot be adapted to the other tasks.  

\subsubsection{Intent Prediction} 
The experimental results of three intent prediction datasets  under the full-data setting are shown in the upper part of Table \ref{tab:intent}. We can see that SPACE-3 achieves the best results on all the datasets.  Specifically, SPACE-3 outperforms the previous SOTA model ConvFiT by 0.78\%, 0.55\% and 1.72\% on BANKING77, CLINC150 and HWU64, respectively. The performance improvements suggest that SPACE-3 has a better ability to discriminate intents via tree-structure based contrastive learning than the general pre-training method in ConvFiT.

\subsubsection{Dialog State Tracking}
To demonstrate whether SPACE-3 can handle multi-turn dialog understanding, we compare it with previous SOTA models on MultiWOZ2.2. As shown in the bottom part of Table \ref{tab:intent}, using SPACE-3 as the base of the TripPy model is able to obtain the best joint goal accuracy in comparison with other baselines. In particular, without any special post-processing or gigantic parameters, SPACE-3 outperforms AG-DST and PLATO-XL (11B parameters) by 0.24\% and 0.34\%, respectively.


\subsubsection{End-to-End Dialog Modeling}
Following previous work \cite{lubis-etal-2020-lava, zhang-etal-2020-probabilistic}, the \texttt{Success} and \texttt{SuccF1} can be used to assess the performance of policy optimization since they indicate whether the system has solved all user's requests and provided correct entities for successful task completion. From Table \ref{tab:multiwoz} we can see that  SPACE-3 achieves best \texttt{Success} or \texttt{SuccF1} on all four benchmarks, and also has promising \texttt{BLEU} on par with other strong baselines. SPACE-3 acquires the best combined scores, outperforming the best baselines by 1.02, 2.18, 3.91 and 4.54 points for CamRest676, In-Car, MultiWOZ2.0 and MultiWOZ2.1, respectively.


\begin{table}[t]
    \caption{Comparison results on four E2E dialog tasks.}
    \label{tab:multiwoz}
    \resizebox{0.47\textwidth}{!}{
    \begin{tabular}{l|cccc|cccc}
    \toprule
    \multirow{2}{*}{Model} & \multicolumn{4}{c|}{CamRest676} & \multicolumn{4}{c}{In-Car} \\ \cline{2-9} 
     & \texttt{Match} & \texttt{SuccF1} & \texttt{BLEU} & \texttt{Comb} & \texttt{Match} & \texttt{SuccF1} & \texttt{BLEU} & \texttt{Comb} \\ \hline
    SEDST & 92.7 & 75.4 & 23.6 & 107.65 & 84.5 & 82.9 & 19.3 & 103.00 \\
    TSCP & 92.7 & 85.4 & 25.3 & 114.35 & 84.5 & 81.1 & 21.9 & 104.70 \\
    LABES & 96.4 & 83.0 & 25.5 & 115.2 & 85.8 & 77.0 & 22.8 & 104.20 \\
    FSDM & 93.5 & 86.2 & \textbf{25.8} & 115.65 & 84.8 & 82.1 & 21.5 & 104.95 \\
    SPACE-3 & \textbf{97.74} & \textbf{88.24} & 23.68 & \textbf{116.67} & \textbf{85.26} & \textbf{83.16} & \textbf{22.92} & \textbf{107.13} \\ \hline\hline
    \multirow{2}{*}{Model} & \multicolumn{4}{c|}{MultiWOZ2.0} & \multicolumn{4}{c}{MultiWOZ2.1} \\ \cline{2-9} 
     & \texttt{Inform} & \texttt{Success} & \texttt{BLEU} & \texttt{Comb} & \texttt{Inform} & \texttt{Success} & \texttt{BLEU} & \texttt{Comb} \\ \hline
    SimpleTOD & 84.40 & 70.10 & 15.01 & 92.26 & 85.00 & 70.50 & 15.23 & 92.98 \\
    DoTS & 86.59 &74.14 &15.06 & 95.43& 86.65& 74.18 &	15.90 & 96.31 \\
    UBAR & \textbf{95.40} & 80.70 & 17.00 & 105.05 & \textbf{95.70} & 81.80 & 16.50 & 105.25 \\
    MTTOD & 90.99 & 82.58 & \textbf{20.25} & 107.04 & 90.99 & 82.08 & 19.68 & 106.22 \\
    SPACE-3 & 95.30 & \textbf{88.00} & 19.30 & \textbf{110.95} & 95.60 & \textbf{86.10} & \textbf{19.91} & \textbf{110.76} \\ \bottomrule
    \end{tabular}
    
    }
\end{table}

\subsection{Ablation Study}
In this section, we attempt to answer three questions regarding SPACE-3: (i) Whether the prompt tokens are useful? (ii) How do the understanding semantic modeling and semantic region modeling contribute to different tasks? (iii) Is the policy semantic modeling crucial for the end-to-end dialog modeling?  

Table \ref{tab:ablation} provides detailed results of our ablation study. We first replace our understanding prompt tokens $p^u$ with the original input representations, calculating all semantic modeling loss based on the representation of \texttt{[CLS]}. As shown in the upper part of Table \ref{tab:ablation}, the performance of the new model (denoted as w/o $p^u$) degrades largely on all understanding tasks, demonstrating the effectiveness of using those prompt tokens. Then, we discard the $L_{slm}, L_{scl}, L_{bow}$ losses individually, and investigate SPACE-3 on the three IP tasks and the DST task. Results show that the span prediction $L_{slm}$ is more important for the intent prediction, while $L_{scl}$ contributes more to dialog state tracking. This may be because multi-turn dialog understanding requires more complex semantics learned by $L_{scl}$. As an auxiliary loss, $L_{bow}$ can improve the model performance slightly. For the E2E task, from the bottom part of Table \ref{tab:ablation} we can find that the results of \texttt{Success} and \texttt{BLUE} significantly deteriorate if we discard  $L_{psm}$, suggesting that dialog policy is the most crucial component for the final  dialog success and overall generation quality since it is the basis for the later dialog generation.


\begin{table}[t]
    \caption{Ablation study on the IP, DST and E2E tasks. Bold numbers indicate the worst performance. The differences are shown in the bracket.}
    \label{tab:ablation}
    \resizebox{0.47\textwidth}{!}{
    \begin{tabular}{l|cccc}
    \toprule
    \multirow{2}{*}{Model} & \multicolumn{3}{c|}{IP} & DST \\\cline{2-5} 
    & BANKING77 & CLINIC150 & HWU64 & \multicolumn{1}{|c}{MultiWOZ2.2} \\ \hline
    SPACE-3 & 94.94 & 97.89 & 94.14 & \multicolumn{1}{|c}{57.50} \\
    \ \ \ \ w/o \texttt{$p^u$} & 94.29(-0.65) & 97.56(-0.33) & 93.49(-0.65) & \multicolumn{1}{|c}{57.15(-0.35)}  \\
    \ \ \ \ w/o $L_{bow}$ & 94.87(-0.07) & 97.76(-0.13) & 93.77(-0.37) & \multicolumn{1}{|c}{57.08(-0.42)} \\
    \ \ \ \ w/o $L_{slm}$ & \textbf{94.12(-0.82)} & \textbf{97.31(-0.58)} & \textbf{92.94(-1.20)} &  \multicolumn{1}{|c}{56.42(-1.08)} \\
    \ \ \ \ w/o $L_{scl}$ & 94.25(-0.69) & \textbf{97.31(-0.58)} & 93.59(-0.55) & \multicolumn{1}{|c}{\textbf{55.96(-1.54)}}  \\
     \hline
    \hline
    \multirow{2}{*}{Model} & \multicolumn{4}{c}{E2E: MultiWOZ2.0} \\ \cline{2-5} 
     & \texttt{Inform} & \texttt{Success} & \texttt{BLEU} & \texttt{Comb} \\ \hline
    SPACE-3 & {95.30} & {88.00} & 19.30 & {110.95} \\
    \ \ \ \ w/o $L_{bow}$ & 94.50(-0.8) & 86.40(-1.6) & 19.01(-0.29) & 109.46(-1.49) \\
    \ \ \ \ w/o $L_{slm}$ & 93.10(-2.2) & 86.10(-1.9) & 18.95(-0.35) & 108.55(-2.40) \\
    \ \ \ \ w/o $L_{scl}$ & \textbf{92.60(-2.7)} & 85.80(-2.2) & 18.70(-0.60) & 107.90(-3.05) \\
    \ \ \ \ w/o $L_{psm}$ & 93.20(-2.1) & \textbf{83.10(-4.9)} & \textbf{18.59(-0.71)} & \textbf{106.74(-4.21)} \\ \bottomrule
    \end{tabular}}
\end{table}

\begin{figure}[t]
    \centering
    \includegraphics[width=0.45\textwidth]{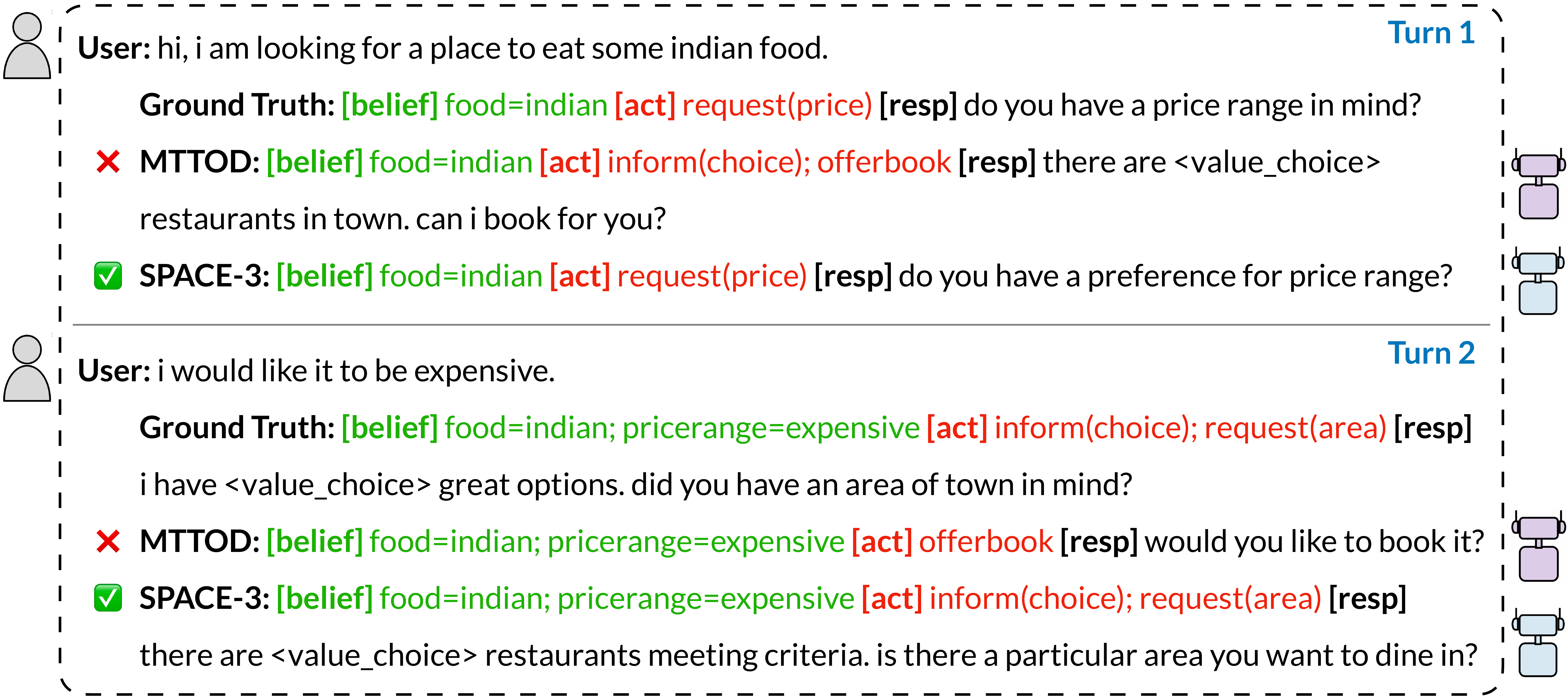}
    \caption{Case Study: Belief states, system acts and delexicalized responses generated by SPACE-3 and MTTOD on MultiWOZ2.0 test data, which are marked in different colors.
    }
    \label{fig:case}
\end{figure}

\subsection{Case Study}
As illustrated in Figure \ref{fig:case}, both SPACE-3 and MTTOD can predict the correct belief states during the conversation, showing the great performance of understanding.
However, MTTOD takes the wrong dialog acts \textit{offerbook} and \textit{inform} and at the beginning turn, and a redundant dialog act \textit{offerbook} at the second turn, which leads to inappropriate responses and a failure for the interaction.
On the contrary, SPACE-3 chooses correct dialog acts for the first two turns so that the whole conversation can steer towards successful task completion, which also verifies our stronger policy planning and generation abilities.

\section{Conclusion}
In this paper, we proposed SPACE-3, a unified pre-trained conversation model, which could be effectively applied to various downstream dialog tasks.
SPACE-3 consisted of four successive components to maintain the task-flow in a task-oriented dialog system via multi-task learning, including a context encoder to encode dialog history, an understanding module to extract structured dialog semantics, a policy module to determine dialog acts, and a generation module to produce responses.
Five dedicated pre-training objectives were employed to optimize the four components of SPACE-3 jointly in a multi-task paradigm.
Experiments demonstrated that SPACE-3 created new SOTA results on several TOD benchmarks,  on both full-data or the few-shot training settings.
We hope that SPACE-3 and the newly collected dialog dataset (\textit{AnPreDial} and \textit{UnPreDial}) can push forward the research in this area. 

\begin{acks}
This work was supported by National Natural Science Foundation of China (No. 61906185), Youth Innovation Promotion Association of CAS China (No. 2020357), Shenzhen Science and Technology Innovation Program (Grant No. KQTD20190929172835662), Shenzhen Basic Research Foundation (No. JCYJ20210324115614039 and No. JCYJ20200109113441941).
This work was supported by Alibaba Group through Alibaba Research Intern Program.
\end{acks}

\bibliographystyle{ACM-Reference-Format}
\bibliography{sample-base}










\end{document}